\documentclass[runningheads]{llncs}
\usepackage[T1]{fontenc}
\usepackage{graphicx}
\usepackage{hyperref}
\usepackage{amsmath,amssymb,amsfonts}
\usepackage{algorithm}
\usepackage{algorithmic}
\usepackage{makecell}
\usepackage{threeparttable}
\usepackage{graphicx}
\usepackage{textcomp}
\usepackage{multirow}
\usepackage{dsfont}
\usepackage{url}
\usepackage{makecell}
\usepackage{threeparttable}
\usepackage{textcomp}
\usepackage{dsfont}
\usepackage[utf8]{inputenc} 
\usepackage[T1]{fontenc}    
\usepackage{hyperref}       
\usepackage{url}            
\usepackage{booktabs}       
\usepackage{amsfonts}       
\usepackage{nicefrac}       
\usepackage{microtype}      
\usepackage{graphicx}
\usepackage{colortbl,booktabs}
\usepackage{graphicx}
\usepackage{epstopdf}
\usepackage{blindtext, graphicx, mathrsfs, amsmath, amsfonts}
\usepackage{booktabs}
\usepackage{multirow, minipage-marginpar}
\usepackage{caption}
\usepackage{subfigure}
\usepackage{color,soul}
\usepackage[textsize=tiny]{todonotes}
\usepackage{algorithm}
\usepackage{algorithmic}
\usepackage{todonotes}
\usepackage{wrapfig,blindtext}
\usepackage{bbm}
\usepackage{wrapfig}
\usepackage{graphicx}
\usepackage{xr}
\begin{document}
%
\title{Brain-Cognition Fingerprinting via Graph-GCCA with Contrastive Learning}
%
\titlerunning{Brain-Cognition Fingerprinting}
\author{Yixin Wang\inst{1} \and 
Wei Peng \inst{2}\and 
Yu Zhang \inst{3} \and
Ehsan Adeli \inst{2} \and \\
Qingyu Zhao\inst{4}\thanks{Corresponding author: \email{qiz4006@med.cornell.edu}} 
\and
Kilian M. Pohl\inst{2}}

%
%
\authorrunning{Y. Wang et al.}

\institute{Department of Bioengineering, Stanford University, Stanford, CA, USA
\and  Dept. of Psychiatry \& Behavioral Sciences, Stanford University, Stanford, CA, USA 
\and Department of Bioengineering, Lehigh University,  Bethlehem, PA, USA
\and Department of Radiology, Weill Cornell Medicine, New York, NY, USA
}
%
%
%
\maketitle              
\vspace{-13pt}
\begin{abstract} 
Many longitudinal neuroimaging studies aim to improve the understanding of brain aging and diseases by studying the dynamic interactions between brain function and cognition. Doing so requires accurate encoding of their multidimensional relationship while accounting for individual variability over time. For this purpose, we propose an unsupervised learning model (called \underline{\textbf{Co}}ntrastive Learning-based \underline{\textbf{Gra}}ph Generalized \underline{\textbf{Ca}}nonical Correlation Analysis (CoGraCa)) that encodes their relationship via Graph Attention Networks and generalized Canonical Correlational Analysis. To create brain-cognition fingerprints reflecting unique neural and cognitive phenotype of each person, the model also relies on individualized and multimodal contrastive learning. We apply CoGraCa to longitudinal dataset of healthy individuals consisting of resting-state functional MRI and cognitive measures acquired at multiple visits for each participant. The generated fingerprints effectively capture significant individual differences and outperform current single-modal and CCA-based multimodal models in identifying sex and age. More importantly, our encoding provides interpretable interactions between those two modalities.

\end{abstract}
%
\section{Introduction}


Longitudinal neuroimaging studies often repeatedly acquire functional MRI and neurocognitive performance measures of study participants to explore the connection between brain function and cognition and their development over time \cite{ji2021mapping,luo2023patterns,lee2023human}. The investigations, however, are often hindered by the lack of computational tools linking such multi-modal, repeated measures.
Despite the advances of machine learning in neuroimaging studies, existing models \cite{zhang2021identification,gao2023multimodal,zhu2018random} are often designed to predict univariate outcome variables, which cannot characterize shared and dissociated neural bases underlying multiple cognitive domains (e.g., working memory, motor functions). Moreover, cross-sectional methods often fail to disentangle the consistent brain functional connectivity of a single subject across multiple visits, from the variations that exist between individuals \cite{finn2016individual,bijsterbosch2017investigations}.
One potential solution to relating multi-dimensional functional and cognitive measures is Canonical Correlation Analysis (CCA) \cite{thompson2000canonical}, which has been successfully applied in a number of cross-sectional studies to identify reproducible brain-cognition mapping. Traditional CCA methods \cite{thompson2000canonical} primarily capture linear associations between modalities, which are not suitable for modeling the complex spatial characteristics inherent in brain connectivity.
An approach that has been highly successful in inferring neural activity patterns in functional MRI are Graph Neural Networks (GNNs) \cite{li2021braingnn,nerrise2023explainable,hu2021gat,peng2022gate}, which have been coupled with the CCA framework to relate two augmented views derived from fMRI signals for spurious factor mitigation \cite{peng2022gate}. Thus, GNN-based CCA might provide a strong foundation for learning brain-cognition mapping, but it remains unclear how such mappings preserve inter-subject variability and intra-subject consistency.

In this work, we propose \underline{\textbf{Co}}ntrastive Learning-based \underline{\textbf{Gra}}ph Generalized \underline{\textbf{Ca}}nonical Correlation Analysis (CoGraCa), aimed at encoding the correlation between brain functional connectivity and cognitive measurements at individual-level while characterizing brain functional differences to create personalized brain-cognition fingerprints that reflect the unique neural and cognitive landscapes of each person. 
We utilize a Graph Attention Network (GAT) to encode brain functional connectivity derived from resting-state functional MRI (rs-fMRI). The GAT is coupled with a generalized CCA (GCCA) \cite{benton2017deep}, which jointly encodes brain function and cognitive scores so that the resulting brain functional networks are aligned with the cognitive data.
To explicitly account for the inter-subject variability and intra-subject consistency, we further design two contrastive learning strategies: i) An individualized contrastive learning approach that regulates the graph embeddings both within and between subjects in the latent space, ensuring that the unique connectivity patterns of each subject are preserved while differentiating between subjects. ii) A longitudinal multimodal contrastive learning that encourages the cross-modal alignment of brain connectivity and cognitive measures across different visits within each subject, maintaining the dynamic evolution of individualized brain-cognition correlation.

Our proposed CoCraCa is cross-validated on a dataset comprising 57 participants, totaling 93 visits containing both fMRI and cognitive measures. The generated ``brain-cognition'' fingerprints demonstrate significant individual differentiation. Validated through downstream sex and age classification task using these fingerprints, CoGraCa achieves higher accuracy scores in comparison to other state-of-the-art single-modal and CCA-based multimodal methods, underscoring its effectiveness in integrating brain connectivity with cognitive data for precise individual characterization. Importantly, CoGraCa enables interpretable correlations between modalities, identifying sex- and age-related functional connectivity and cognitive measures that align with established neuroscience research.
\vspace{-8pt}
\begin{figure}[t]
\centering
\includegraphics[width=0.9\textwidth]{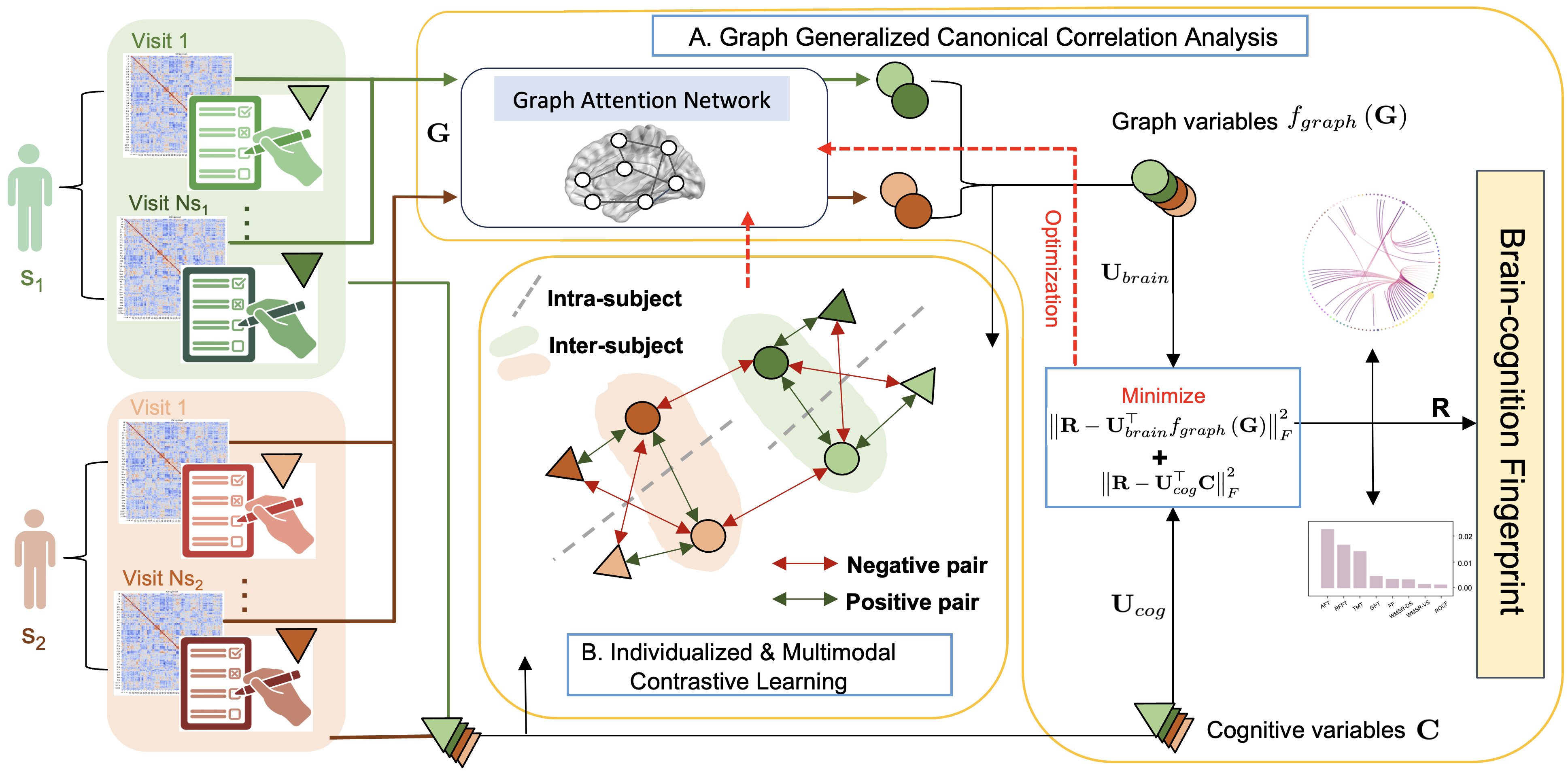}
\caption{Overview of our model. (A) Brain functional connectivity of each subject is encoded using graph attention networks (GAT) into graph embedding. The learned embedding (graph variables) and cognitive measures (cognitive variables) are mapped to a shared ``brain-cognition'' space via generalized canonical correlation analysis. The GAT weights are updated by optimizing the maximum correlation between two modalities. (B) An individualized contrastive learning differentiates inter-subjects brain connectivity and a multimodal contrastive learning to align brain connectivity and cognitive measures across multiple visits within subjects, capturing intra-subject and cross-modal dynamics.}\label{fig:network}
\vspace{-14pt}
\end{figure}
\vspace{-7pt}
\section{Method}
Let $S$ be the number of subjects and $N$ be the number of visits across all subjects, with $N_s$ representing the number of visits for subject $s$. Each visit $i$ contains a set of cognitive measures $\mathcal{C}_i$ and a connectivity matrix encoding the Pearson correlation between the fMRI signal of brain Regions of Interests (ROIs). The connectivity matrix is represented as a graph $\mathcal{G}_i=\left(\mathbf{A}_{\mathcal{G}_i}, \mathbf{D}_{\mathcal{G}_i}\right)$ consisting of $V$ nodes (representing the ROIs). $\mathbf{A}_{\mathcal{G}_i} \in \mathbb{R}^{V \times V} $ is the adjacent matrix consisting only of positive correlations (anticorrelations are set to 0) \cite{weissenbacher2009correlations,li2022joint}. $\mathbf{D}_{\mathcal{G}_i} \in \mathbb{R}^{V \times D}$ is the attribute matrix, represented by each ROI's ``connection profile'' of length $D$, as defined in \cite{cui2022braingb}. Based on the set of pairs $\left\{\mathbf{G}, \mathbf{C}\right\}=\left\{\left(\mathcal{G}_1, \mathcal{C}_1\right),\left(\mathcal{G}_2, \mathcal{C}_2\right), \cdots,\left(\mathcal{G}_n, \mathcal{C}_n\right)\right\}$, the goal of our approach is to learn a brain-cognition representation (or ``Brain-Cognition fingerprints'') $\mathbf{R} = \left\{\mathcal{R}_1, \mathcal{R}_2, \cdots, \mathcal{R}_n\right\}$. We determine the optimal $\mathbf{R}$ by regularizing Graph GCCA (Fig.\ref{fig:network} A) by individualized and multimodal contrastive learning (Fig.\ref{fig:network} B). We now describe these components in further detail.

\noindent\textbf{Graph Generalized Canonical Correlation Analysis. }
For each input connectivity graph $\mathcal{G}_i$, we adopt Graph Attention Layers (GAT) \cite{velickovic2017graph} to learn its encoding into a node embedding $\boldsymbol{h}_{\mathcal{G}_i} \in  \mathbb{R}^{V \times r}$, where the embedding $\boldsymbol{h}_p \in  \mathbb{R}^{1 \times r}$ of node $p$ is updated by aggregating the features of its 1-hop neighborhood nodes $\mathcal{N}_p$ through self-attention mechanism. Specifically, the attention between $p$ and its neighboring node $q$ at layer $k$ is calculated by:
$
a_{p q}^k\left(\boldsymbol{h}_p^k, \boldsymbol{h}_q^k\right)=\operatorname{Softmax}\left(\sigma\left(m^T\left[\mathrm{W} \boldsymbol{h}_p^k \| \mathrm{W} \boldsymbol{h}_q^k\right]\right)\right),
$
where $\|$ denotes a concatenation operation, $\sigma$ is a non-linear activation function ReLU, $m$ is a trainable single-layer feed-forward neural network, and $\mathrm{W}$ is a trainable weight matrix. The node representation at layer $k+1$ will be further obtained by:
$
\boldsymbol{h}_p^{k+1}= \sigma\left(\sum_{q \in \mathcal{N}_p} a_{p q}^k \mathrm{W} \boldsymbol{h}_q^k\right).
$
The node embeddings $\left\{\boldsymbol{h}_p\right\}_{p \in V}$ of each $\mathcal{G}_i$ from the last layer are further applied to a global mean pooling operation to obtain a set of graph variables $\boldsymbol{h}_{\mathcal{G}_i}$. We aim to learn this encoding function $\mathcal{H} := f_{graph}\left(\mathbf{G}\right)$ through maximizing its correlation with cognitive measures.
Specifically, the representations $\mathcal{H} := \left\{\boldsymbol{h}_{\mathcal{G}_i}\right\}_{\mathcal{G}_i \in \mathbf{G}}$ learned from the complex brain connectivity, along with the cognitive measures, are treated as two sets of canonical variables that will be projected into a shared ``brain-cognition'' space to obtain $\mathbf{R}$ through GCCA\cite{GCCA,benton2017deep}.
Note, the cognitive measures are not subjected to any encoder to ensure its direct guidance to fMRI generation and integration.

The unsupervised CCA optimization is expressed as maximizing the sum of correlations between $\mathbf{R}$ and each modality, defined by the loss function:
$
\mathcal{L}_{\text {corr}} = \left\|\mathbf{R}-\mathbf{U}_{brain}^{\top} f_{graph}\left(\mathbf{G}\right)\right\|_F^2 + \left\|\mathbf{R}-\mathbf{U}_{cog}^{\top} \mathbf{C}\right\|_F^2, 
\text { s.t. } \mathbf{R} \mathbf{R}^{\top}=\mathbf{I} .
$
$\mathbf{U}_{brain}$ and $\mathbf{U}_{cog}$ are linear transformation matrix (canonical loadings) of the two variables from brain connectivity and cognition measures. Note, $\mathbf{R}$ is the shared representation in the ``brain-cognition'' space, which is obtained by solving an eigenvalue problem. Following \cite{benton2017deep}, for the sets of brain graphs $\mathbf{G}$, we define the covariance
matrix as $\mathbf{Cov}=f_{graph}\left(\mathbf{G}\right) f_{graph}\left(\mathbf{G}\right)^{\top}$ and obtain the positive semi-definite matrix $\mathbf{P}_{brain}=f_{graph} \left(\mathbf{G}\right)^{\top} \mathbf{Cov}^{-1} 
 f_{graph}\left(\mathbf{G}\right)$. Similarly, we obtain $\mathbf{P}_{cog}$ from $\mathbf{C}$ and we stack them as $\mathbf{M}=\mathbf{P}_{brain}+\mathbf{P}_{cog}$. Then, the eigenvectors of $\mathbf{M}$ will be constructed into $\mathbf{R}$ which maximally 
 and linearly correlates non-linear transformations of brain functional connectivity and cognition measures.
This optimization problem is solved by estimating the gradient of the objective on samples that are mapped through $f_{graph}$ and using back-propagation to update weights within $f_{graph}$.

\noindent\textbf{Individualized Contrastive Learning.}
To capture the inherent individual variability present in brain functional connectivity, we design an individualized contrastive learning strategy where pairs of brain connectivity are constructed from all $N$ visits across $S$ subjects. Pairs from the same subject $s$ are considered to be similar and thus are labeled as positive pairs (see Fig. \ref{fig:network}B). Conversely, pairs from different individuals are likely to be dissimilar and are labeled negative accordingly. Specifically, given the sets of graph embeddings $\mathcal{H}: = \left\{\boldsymbol{h}_{\mathcal{G}_i}\right\}_{\mathcal{G}_i \in \mathbf{G}}$, we define a subject indicator $\mathbbm{1}_{\mathbf{y}_{i j}=1}$, where $\mathbf{y}_{i j}=1$ denotes a pair of graph embeddings $\boldsymbol{h}_{\mathcal{G}_i}$ and $\boldsymbol{h}_{\mathcal{G}_j}$ are from the same subject, otherwise, $\mathbf{y}_{i j}=0$. The individualized contrastive loss can then be achieved by \\
$
\mathcal{L}_{\text {ind}}=-\frac{1}{N}\sum_{i \in N}\sum_{j \in N} \mathbbm{1}_{\mathbf{y}_{i j}=1} \log \frac{\exp \left(\operatorname{sim}\left(\boldsymbol{h}_{\mathcal{G}_i}, \boldsymbol{h}_{\mathcal{G}_j}\right) / \tau\right)}{\sum_{k \in N, k \neq i} \exp \left(\operatorname{sim}\left(\boldsymbol{h}_{\mathcal{G}_i}, \boldsymbol{h}_{\mathcal{G}_k}\right) / \tau\right)},
$
where $\operatorname{sim}(\cdot)$ denotes the cosine similarity and $\tau > 0$ is a temperature parameter that controls the separation of subjects.
By doing so, graph embeddings from the same subject $s$ are pulled closer together than embeddings from different subjects.

\noindent\textbf{Multimodal Contrastive Learning.}
Meanwhile, we have to ensure the above individual-level separation keeps the longitudinal differences within each subject. To achieve this goal, we apply a multimodal contrastive learning, inspired by CLIP \cite{radford2021learning}, yet specifically tailored to within-subject pairs for our multimodal features, i.e., brain connectivity and cognitive measures, to capture this intra-subject and cross-modal dynamics. 
For subject $s \in S$ with the number of visits $N_s > 1$, given the paired brain graph $\mathcal{G}_i^s$ and cognitive measures $\mathcal{C}_i^s$ across $N_s$ visits, we aim to maximize their similarity from the same visit and minimize the similarity from different visits within each subject. This multimodal contrastive learning is achieved by:
$
\mathcal{L}_{\text {mul}}=-\frac{1}{S}\sum_{s \in S} \sum_{i \in N_s} \log \frac{\exp \left(\operatorname{sim} \left(\boldsymbol{h}_{\mathcal{G}_i^s}, \mathcal{C}_i^s\right) / \tau\right)}{\sum_{k \in N_s, k \neq i} \exp \left(\operatorname{sim} \left(\boldsymbol{h}_{\mathcal{G}_i^s}, \mathcal{C}_k^s\right) / \tau\right)}.
$

By accounting for both the brain connectivity variability between subjects and variability in correlation with cognitive measures within each subject, the model can derive a more individualized representation that integrates the longitudinal variations in brain connectivity specific to cognitive measures. The final objective function is defined as $\mathcal{L}_{\text {total }}:=\mathcal{L}_{\text {corr}} + \lambda_1 \mathcal{L}_{\text {ind}} + \lambda_2 \mathcal{L}_{\text {mul}}$, where $\lambda_1$ and $\lambda_2$
are trade-off parameters to balance the two contrastive learning procedures. 

%
%
\vspace{-8pt}
\section{Experimental Results}
\vspace{-5pt}
\noindent\textbf{Dataset.}
\label{sec3.1}
Our study utilizes the SRI dataset (PIs: Pfefferbaum and Sullivan) consisting of rs-fMRI (3T GE Discovery MR750 scanner, 8-channel head coil, echo time=30ms, dwell-time=0.388ms, TR=2200ms, 2.5mm isotropic after upsampling) of 417 subjects (822 visits). Of them, 195 participants (275 visits) completed cognitive testing at the same visit as the rs-fMRI was acquired. The cognitive measurements are summarized in 16 domain-specific scores. Of the 195 subjects, 57 subjects (89 visits, age: 58.53±10.57 years) are normal controls with 21 females (33 visits) and 36 males (56 visits). Of each of the 89 rs-fMRI, the pipeline by \cite{honnorat2022alcohol} extracts connectivity matrix across 111 ROIs. Each entry in that matrix is the Pearson correlation between the rs-fMRI signals of two ROIs. 

\noindent\textbf{Implementation Details.}
We implement the proposed model CoGraCa using PyTorch with the Adam optimizer and a learning rate of 0.001. Our graph encoder is composed of two GAT layers, with hidden units=32. The dimension of the node embedding $r$ is set as 16 and the number of canonical variants (i.e., dimension of $\mathbf{R}$) are set as 16. $\tau$ is set as 0.9 and $\lambda_1$ and $\lambda_2$ are set as 1.5 and 0.5, respectively. The model is trained for 1000 epochs using five-fold cross-validation with folds defined by subjects to ensure that all visits from a single subject are assigned to the same fold. For each test fold, the model is trained on the remaining data to optimize the model's parameters and obtain canonical loadings $\mathbf{U}_{brain}, \mathbf{U}_{cog}$. After training is completed, the ``Brain-Cognition'' representation  $\mathbf{R}$ is generated for each sample from the test fold. Codes will be made available at \hyperref[https://github.com/Wangyixinxin/BrainCog]{https://github.com/Wangyixinxin/BrainCog}. 

\subsection{Individual Variability of ``Brain-Cognition'' Fingerprints}

We assess whether our derived representations could capture individual-specific features more effectively than other CCA-based methods. 
\begin{figure}[t]
\includegraphics[width=0.95\textwidth]{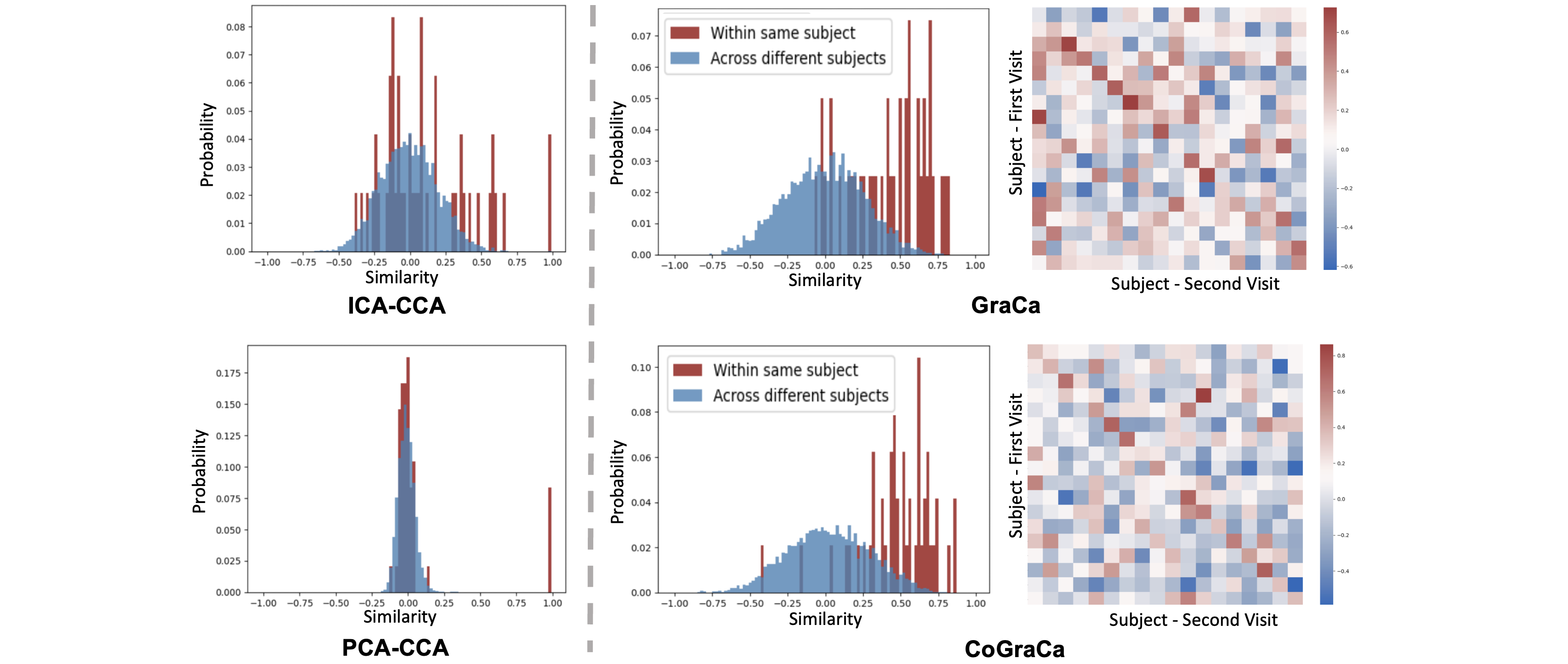}
\centering
\caption{Histograms show the similarity in the representations between visits within subject (intra-subject, red) vs. across subjects (inter-subject, blue). Of all methods, CoGraCa model produced the most individualized representation, i.e., the intra-subject similarity is relatively high compared to the inter-subject similarity. 
 }
\label{individual_Ver}
\end{figure}

\noindent\textbf{Experimental Setup.} For comparison, we repeat the five fold cross-validation by applying Principal Component Analysis (PCA) and Independent Component Analysis (ICA) to the connectivity matrices before performing CCA analysis. PCA reduces the connectivity matrices to 544 independent components (which accounted for 95\% of the data variance) and ICA to 20 independent components (chosen based on the best performance compared with 10,15,20,25 components). Separately for PCA and ICA, the components are fed into CCA in conjunction with cognitive measures to obtain a representation, labelled as PCA-CCA and ICA-CCA in our comparison. Finally, we run CoGraCa omitting contrastive learning (referred to as  GraCa). For each model, we compute the similarity between each pair of representations both within subjects across different visits and between subjects across their visits using Pearson correlation.

\noindent\textbf{Results.} 
The histograms in Fig.\ref{individual_Ver} for ICA-CCA and PCA-CCA show that the distributions of within-subject similarity (red) and between-subject similarity (blue) largely overlap, indicating a lack of individual distinctiveness in the representations generated by these models. The two distributions are much better separated by the representations generated by GraCa and CoGraCa, with individuals being significantly distinct from each other (Mann-Whitney U test, CoGraCa: p-value < 0.0001, GraCa: p-value < 0.0001). CoGraCa is also associated with a larger Wasserstein distance (0.45), i.e., larger distribution separation, compared with GraCa (0.39).
We also measure the similarity between visits from the same and diﬀerent subjects for 18 subjects with two visits in Fig.\ref{individual_Ver} (See GraCa and CoGraCa). 
The correlation matrices reveal that CoGraCa produces highly differentiable individualized representations, where each participant exhibits a high correlation with their own across visits (as reflected in the diagonal of correlation matrix) and low correlations from visits from other cohorts, leading to individualized ``brain-cognition'' fingerprints. 
GraCa yields highly similar representations within individual participants but also displays a higher degree of similarity to other subjects than CoGraCa.
%
%
\vspace{-5pt}
\subsection{Validating Fingerprints with Downstream Tasks}
\vspace{-4pt}
We conduct a quantitative analysis to determine if the integrated ``brain-cognition'' representations from CoGraCa enhance the accuracy of identifying specific individual characteristics compared to GraCa and other CCA-based multimodal methods that aim to correlate brain function and cognition. Based on the trained model for each of the 5 test folds, the representations for both the training and testing sets are generated for downstream tasks without any additional fine-tuning of the model. 

The downstream tasks focus on predicting age (older vs younger) and sex as brain function and cognition often differentiate between these cohorts \cite{bachmann2023age,tomasi2023measures}. Given the relatively small sample size of our data set, the task of age prediction is confined to younger ($\leq$60 years, 47 visits, 38.3$\%$ are females ) versus older (>60 years, 42 visits, 35.7$\%$ are females) as in \cite{statsenko2021predicting}. The sex ratio is similar between those two cohorts according to Chi-square test (p=0.97). For identifying sex, males (age: 59.21±10.01 years) and females (age: 57.38±11.68 years) have similar age (t-test, p=0.45).  The classification model is a multi-layer perceptron (MLP) containing two fully-connected layers of dimension 64 and 32 with ELU and a dropout rate of 0.5. Due to the limitation of the small data size, we repeat the cross-validation of the MLP 10 times using different seed points for initialization. For each cross-validation, we record the balanced accuracy (BACC).  

\begin{table}[t]
    \centering
     \caption{Balanced accuracies on sex and age prediction tasks. 
     Results were averaged across 5 folds and run 10 times with random seeds. The best results are shown in \textbf{bold}.}
     \setlength{\tabcolsep}{0.8mm}{
    \begin{tabular}{llccc}
\toprule[1pt]
& & \multicolumn{1}{c}{Sex Classification} & \multicolumn{1}{c}{Age Classification} \\ \hline 
\multicolumn{2}{l}{Functional MRI-only} \\
~~~~& ICA &  56.05$\pm$2.3  & 69.38$\pm$2.24 \\
& Vanilla GCN \cite{kipf2016semi} &54.73$\pm$2.43&58.75$\pm$2.57 \\ 
& BrainGNN \cite{li2021braingnn} &61.25$\pm$2.12&63.57$\pm$1.09 \\ 
& GAT-LI \cite{hu2021gat} & 58.22$\pm$1.59 &  62.73$\pm$1.99 \\ \hline
\multicolumn{2}{l}{Cognition-only} & 73.45$\pm$1.24 &  71.81$\pm$2.01 \\  \hline
\multicolumn{2}{l}{multimodal} \\
& PCA-CCA  & 72.93$\pm$1.56 & 71.31$\pm$2.01 \\
& ICA-CCA & 65.42$\pm$ 2.27 & 63.31$\pm$2.14 \\
& SDGCCA \cite{moon2022sdgcca} & 69.11$\pm$1.94 & 65.7$\pm$2.57 \\
& GraCa & 74.14$\pm$1.46 & 70.30$\pm$1.52  \\
& CoGraCa & \textbf{74.90$\pm$1.78} & \textbf{73.26$\pm$1.21}\\
\bottomrule[1pt]
\end{tabular}}
    \label{results}
    
 \vspace{-8pt}
\end{table}

\noindent\textbf{Baseline.} 
We compare our method against single modality-based (fMRI-only and cognition-only) and multimodal CCA-based approaches. With respect to fMRI-only, we derive representations via ICA (20 components) and perform classification with MLP. Supervised methods include vanilla GCN \cite{kipf2016semi}, BrainGNN \cite{li2021braingnn}, and GAT-LI \cite{hu2021gat}. Multimodal methods were PCA-CCA and ICA-CCA. We also employ the state-of-the-art supervised SDGCCA approach \cite{moon2022sdgcca}. Each method is trained and tested using the same experimental setup as for GraCa and CoGraCa. 

\noindent\textbf{Results.} 
Table \ref{results} lists the average and standard deviation accuracy scores across the 10 runs. Solely relying on correlation matrices (i.e., fMRI) results in relatively low scores regardless of representation due to the low signal-to-noise ratio of that modality.  The accuracy scores of the multimodal baseline methods (PCA-CCA, ICA-CCA, and SDGCCA) are higher but also lower than only relying on the cognitive measures, which suggests that they are not able to properly integrate the multimodal data. 
Our method achieves that goal and has the highest BACC for sex and age. All its accuracy scores are higher than GraCa, which aligns with our expectation that multimodal contrastive learning effectively maintains the longitudinal intra-individual distinctions.

\subsection{Functional Connectivity and Cognition Interpretation}
A significant achievement of our approach is the extraction of cognitive-related functional connectivities and identifying the most relevant features from both modalities linked to sex or age distinctions. Using SHapley Additive exPlanations (SHAP) \cite{sundararajan2020many}, we identify key features in ``brain-cognition'' representations that drive predictions. These features correspond to specific CCA components, leading us to extract canonical loadings $\mathbf{U}_{brain}$ and $\mathbf{U}_{cog}$ that highlight the significance of graph and cognitive variables.
\label{sec3.2}
\begin{figure}[t]
\centering
\includegraphics[width=\textwidth]{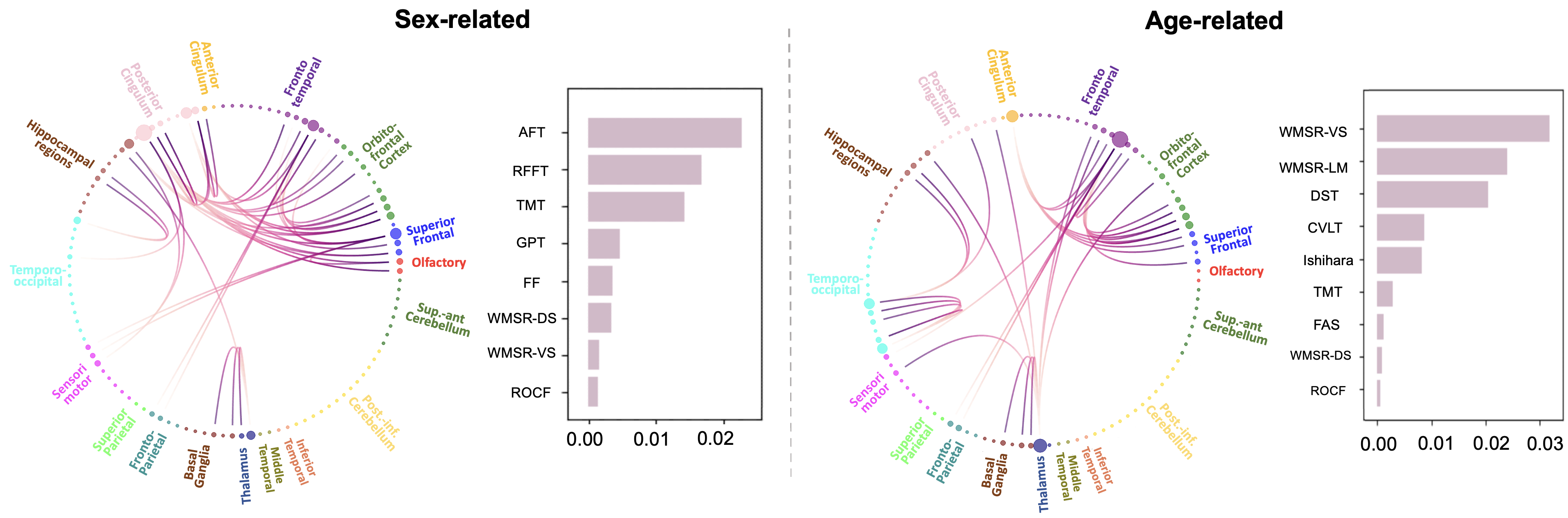}
\caption{Identified functional connectivity and rank-ordered positive loadings of cognitive variables of task-related CCA component, in line with \cite{zhang2018functional,weis2019sex,ramanoel2019age,hsieh2023age,morrison1979reliability,margolis1984age}. Functional regions and cognitive variables were detailed in Supplemental Fig.\ref{supplement-s6}.} 
\label{contribution}
\vspace{-11pt}
\end{figure}
By combining graph variables with the attention matrices obtained from GAT that indicate the learned connectivity from CoGraCa, we derive the functional connectivity pattern identifying sex and age (Fig. \ref{contribution}). Color-coding of brain functional regions is defined according to \cite{honnorat2022alcohol} and their contribution is encoded by the node size. With respect to sex, our method identifies dense connectivity within \textcolor[rgb]{0,0.5,0}{Orbito-frontal Cortex (OFC)} , \textcolor[rgb]{0.5,0,0.5}{Frontotemporal (FT)}, \textcolor[rgb]{1,0.8,0.8}{Posterior Cingulum (PCC)}, and \textcolor[rgb]{0.4,0.2,0.2}{Hippocampal (HPC) regions},  which is in line with the literature \cite{zhang2018functional,weis2019sex}. For age distinctions, significant connections involve \textcolor[rgb]{0.2,1,1}{Temporo-occipital (TO)}, \textcolor[rgb]{0.4,0.2,0.2}{HPC}, \textcolor[rgb]{0,0.5,0}{OFC}, and \textcolor[rgb]{0.2,0.2,0.8}{Superior Frontal (SF)} regions, resonating with neuroscience findings on age-related neural alterations on Parahippocampal, Occipital and Prefrontal areas \cite{ramanoel2019age,hsieh2023age}. See Supplemental Fig.\ref{supplement-s1} and Fig.\ref{supplement-s2} for additional insights on the robustness of connectivities across folds and Fig.\ref{supplement-s3} for connectivity patterns from other CCA components.
These brain functional patterns interacting with cognitive measures are in line with the literature: Alternate Finger Tapping Test (AFT) is important for identifying sex \cite{morrison1979reliability} and, for age, the Wechsler Memory Scale-Revised Test (WMSR)\cite{margolis1984age}, which assesses visual/logical memory. Interestingly, WMSR is correlated correctly with functional regions related to memory (e.g. \textcolor[rgb]{0.2,1,1}{TO} and \textcolor[rgb]{0.4,0.2,0.2}{HPC}) revealing that CoCraCa provides a meaningful integration between brain function and cognition. 
\vspace{-8.7pt}
\section{Conclusion}
\vspace{-8.5pt}
In this work, we introduced a novel unsupervised approach, CoGraCa, to accurately encode brain function coupled with cognition as captured by longitudinal rs-fMRI and cognitive testing. CoGraCa generates ``brain-cognition'' fingerprints capturing the unique neural and cognitive landscapes of individuals across time by coupling Graph GCCA with individualized and multimodal contrastive learning. We measure the accuracy of CoGraCa by using the encoding to identify the sex and age in individuals. Our multimodal encoding has a higher balanced accuracy than several state-of-the-art representations. More importantly, CoGraCa allows us to identify the brain-cognition relationship important for these tasks.

\subsubsection{Acknowledgments} The work was partly funded by the National Institute of Health (DA057567,  AA05965, AA017347, AA010723, MH129694, MH130956, AG080425, AA028840), the DGIST Joint Research Project, the 2024 Stanford HAI Hoffman-Yee Grant, the Stanford HAI-Google Cloud Credits Award, BBRF Young Investigator Grant and the Lehigh University FIG (FIGAWD35) and CORE (001250) grants.


%
%
%
\bibliographystyle{splncs04}
\bibliography{ref}

\title{Supplement of \\
Brain-Cognition Fingerprinting via Graph-GCCA with Contrastive Learning}
\titlerunning{Brain-Cognition Fingerprinting}
\author{Yixin Wang\inst{1} \and 
Wei Peng \inst{2}\and 
Yu Zhang \inst{3} \and
Ehsan Adeli \inst{2} \and \\
Qingyu Zhao\inst{4}\thanks{Corresponding author: \email{qiz4006@med.cornell.edu}} 
\and
Kilian M. Pohl\inst{2}}

\institute{Department of Bioengineering, Stanford University, Stanford, CA, USA
\and  Dept. of Psychiatry \& Behavioral Sciences, Stanford University, Stanford, CA, USA 
\and Department of Bioengineering, Lehigh University,  Bethlehem, PA, USA
\and Department of Radiology, Weill Cornell Medicine, New York, NY, USA
}

\maketitle  
 
\setcounter{figure}{0}
\renewcommand{\thefigure}{S\arabic{figure}}
 \vspace{-10pt}
\begin{figure}
\includegraphics[width=\textwidth]{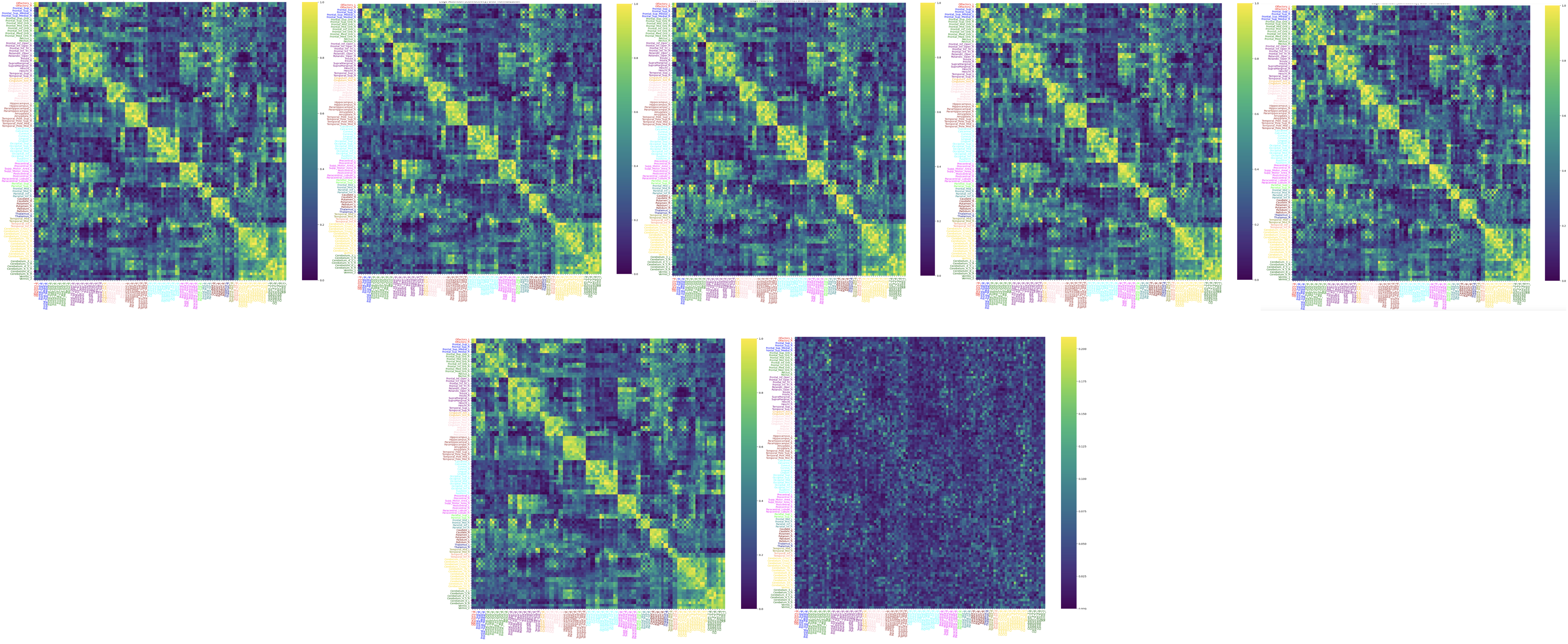}
\caption{Visualization of attention weight matrices learned from 5 folds in CoGraCa (first row). The averaged attention weight matrices and their standard deviations (second row) demonstrate the model's consistent attention patterns. 
 }\label{supplement-s1}
 \vspace{-10pt}
\end{figure}
\vspace{-10pt}
\begin{figure}
\includegraphics[width=\textwidth]{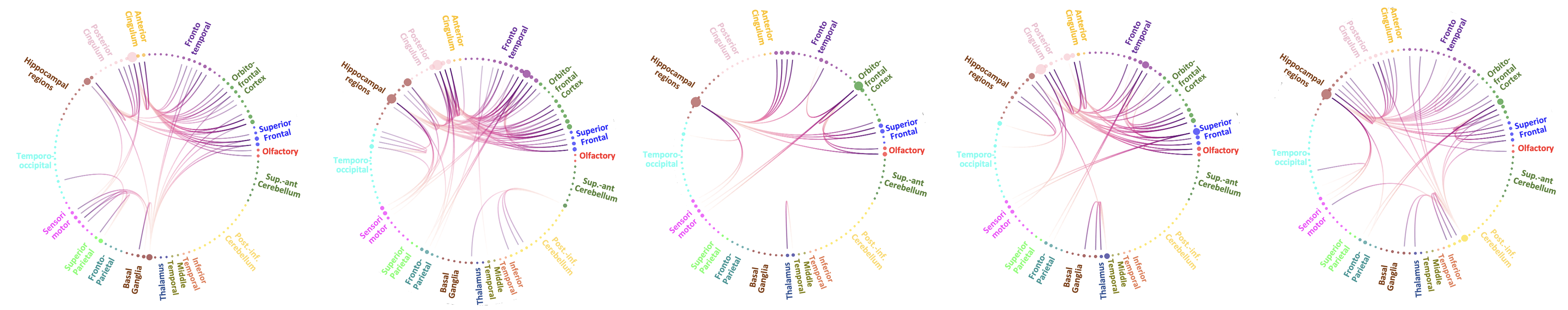}
\caption{Functional connectivity of Top 1 sex-related CCA component from five folds. They revealed similar dense connectivity patterns within \textcolor[rgb]{0,0.5,0}{Orbito-frontal Cortex (OFC)} , \textcolor[rgb]{0.5,0,0.5}{Frontotemporal (FT)}, \textcolor[rgb]{1,0.8,0.8}{Posterior Cingulum (PCC)}, and \textcolor[rgb]{0.4,0.2,0.2}{Hippocampal (HPC)} regions.
 }\label{supplement-s2}
 \vspace{-200pt}
\end{figure}
 \vspace{-20pt}
\begin{figure}
\centering
\includegraphics[width=0.8\textwidth]{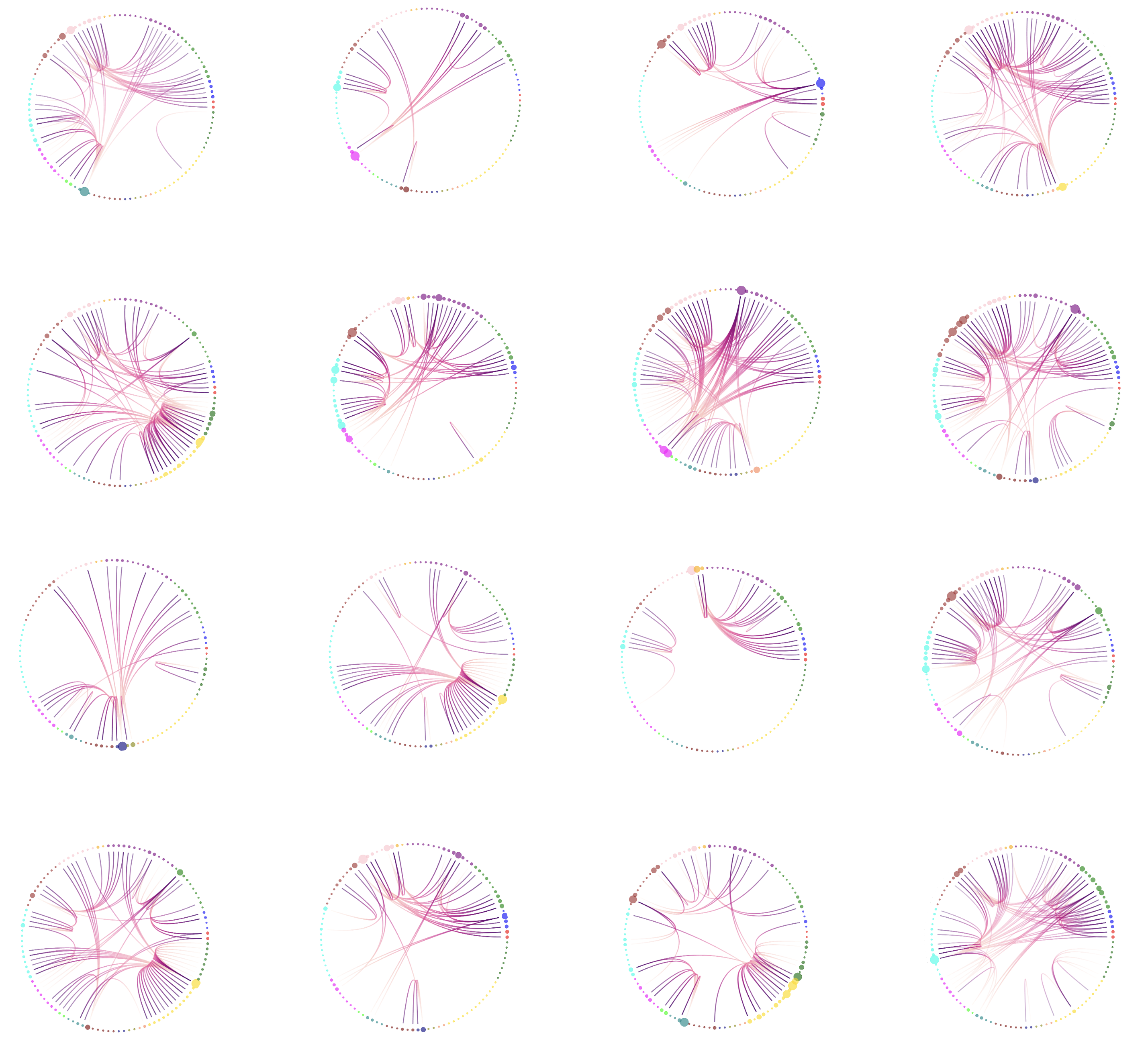}
\caption{Functional connectivity of all 16 CCA components that represent different brain functional connectivity in one fold.
 }\label{supplement-s3}
  \vspace{-20pt}
\end{figure}
 \vspace{-20pt}
\begin{figure}
\centering
\includegraphics[width=\textwidth]{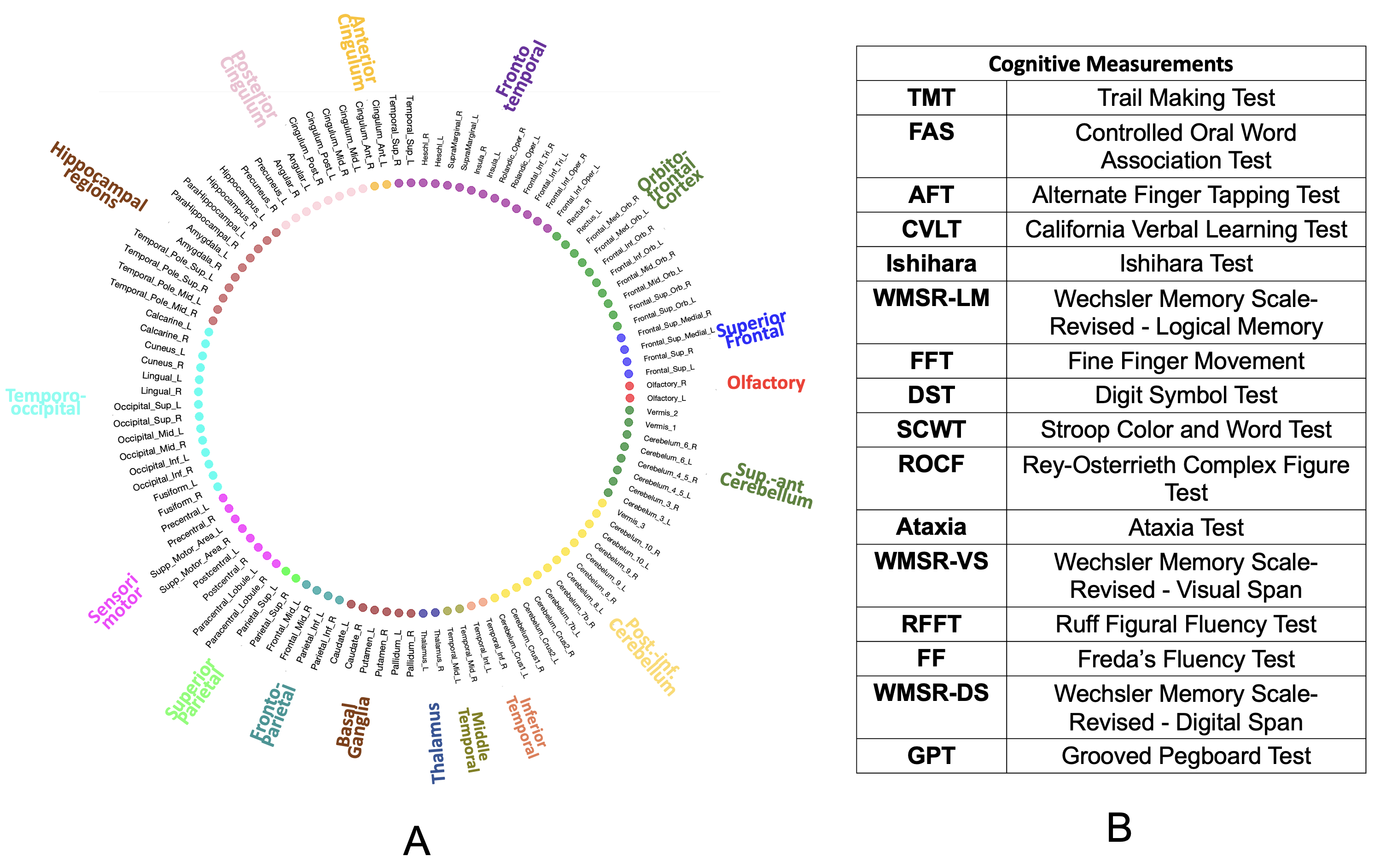}
\caption{(A) 17 functional areas with brain ROI labeled. (B) 16 cognitive measurements.
 }\label{supplement-s6}
  \vspace{-10pt}
\end{figure}

\end{document}